\documentclass[final]{cvpr}

\usepackage{times}
\usepackage{epsfig}
\usepackage{graphicx}
\usepackage{amsmath}
\usepackage{amssymb}

\usepackage[switch]{lineno}
\usepackage{bm}
\usepackage{amssymb}
\usepackage{amsmath}
\usepackage{multirow}
\usepackage{algorithmic}
\usepackage[ruled]{algorithm2e}
\usepackage{times}
\usepackage{epsfig}
\usepackage{graphicx}
\usepackage{amsmath}
\usepackage{amssymb}
\usepackage{subfigure}
\usepackage{makecell}
\usepackage{booktabs}
\usepackage [latin1]{inputenc}

\usepackage[pagebackref=true,breaklinks=true,colorlinks,bookmarks=false]{hyperref}

\def\cvprPaperID{62} 
\def\confYear{CVPR 2021}

\begin{document}

\title{FocusedDropout for Convolutional Neural Network}

\author{\textbf{Tianshu Xie}\textsuperscript{1}\\
{\tt\small xietianshucs@163.com}
\and
\textbf{Minghui Liu}\textsuperscript{1}\\
{\tt\small minghuiliuuestc@163.com}
\and
\textbf{Jiali Deng}\textsuperscript{1}\\
{\tt\small julia\_d@163.com}
\\
\\
\and
\textbf{~~~Xuan Cheng}\textsuperscript{1}\\
{\tt\small ~cs\_xuancheng@163.com}
\and
\textbf{~~~~~~Xiaomin Wang}\textsuperscript{1}\\
{\tt\small  ~~xmwang@uestc.edu.cn}
\and
\textbf{~~Ming Liu}\textsuperscript{1*}\\
{\tt\small csmliu@uestc.edu.cn}
}

\maketitle

\begin{abstract}
In convolutional neural network (CNN), dropout cannot work well because dropped information is not entirely obscured in convolutional layers where features are correlated spatially. Except randomly discarding regions or channels, many approaches try to overcome this defect by dropping influential units. In this paper, we propose a non-random dropout method named FocusedDropout, aiming to make the network focus more on the target. In FocusedDropout, we use a simple but effective way to search for the target-related features, retain these features and discard others, which is contrary to the existing methods. We found that this novel method can improve network performance by making the network more target-focused. Besides, increasing the weight decay while using FocusedDropout can avoid the overfitting and increase accuracy. Experimental results show that even a slight cost, 10\% of batches employing FocusedDropout, can produce a nice performance boost over the baselines on multiple datasets of classification, including CIFAR10, CIFAR100, Tiny Imagenet, and has a good versatility for different CNN models.
\end{abstract}

\section{Introduction}

In recent years, deep neural networks have made significant achievements for many computer vision tasks such as image classification, detection, and segmentation, benefit from deep layers and millions of neurons. It also leads to inadequate training of a part of units in the networks. Dropout\cite{srivastava2014dropout} is proposed as a widely-used regularization method to fight against the overfitting, which stochastically sets to zero the activations of hidden units during training. For deep CNN, dropout works well in fully-connected layers, but its effect is still not apparent in convolutional layers, where features are correlated spatially. When the features are strongly correlated between adjacent neurons, the information of discarded neurons cannot be completely obscured.

\begin{figure}[t]
\begin{center}
\includegraphics[width=0.95\linewidth]{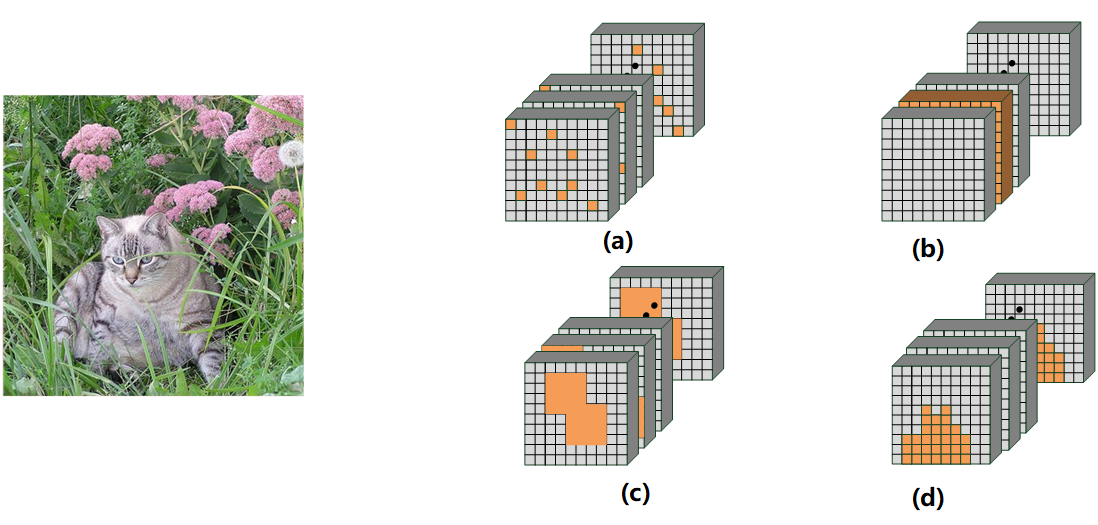}
\caption{Four forms of dropout on CNN. (a) Dropout randomly drops units on each channel. But it doesn¡¯t work well on CNN. (b) SpatialDropout randomly drops channel instead of units.(c) DropBlock randomly drops some contiguous regions inspired by Cutout. (d) Some non-random dropout methods discard influential units. The orange neurons, channels, contiguous regions present the discarded parts, and the gray will be retained.}
\label{fig1}
\end{center}
\end{figure}

Many researchers have observed this defect and tried to improve the dropout method to regularize the convolutional network better. As shown in Figure \ref{fig1}, SpatialDropout\cite{tompson2015efficient} randomly discards entire channels from whole feature maps. DropBlock\cite{ghiasi2018dropblock} randomly discards units in a contiguous region of a channel instead of substantive units. Guided dropout\cite{keshari2019guided}, AttentionDrop\cite{ouyang2019attentiondrop}, and CamDrop\cite{wang2019camdrop} search the influential units in the network through their ways and drop them to enhance the generalization performance of the network.

From another perspective, the foreground of the image is the main focus of the classification task, which means accurately finding the foreground is equivalent to solving the problem of image classification. This enlightened us on using dropout to enhance the network's ability to recognize the classified objects by keeping the foreground related units in training and dropping other units, which is contrary to the existing methods that drop influential ones. There are two challenges: one is how to identify the foreground related units, and how can FocusedDropout still prevent overfitting.

\begin{figure*}[t]
\begin{center}
\includegraphics[width=0.9\linewidth]{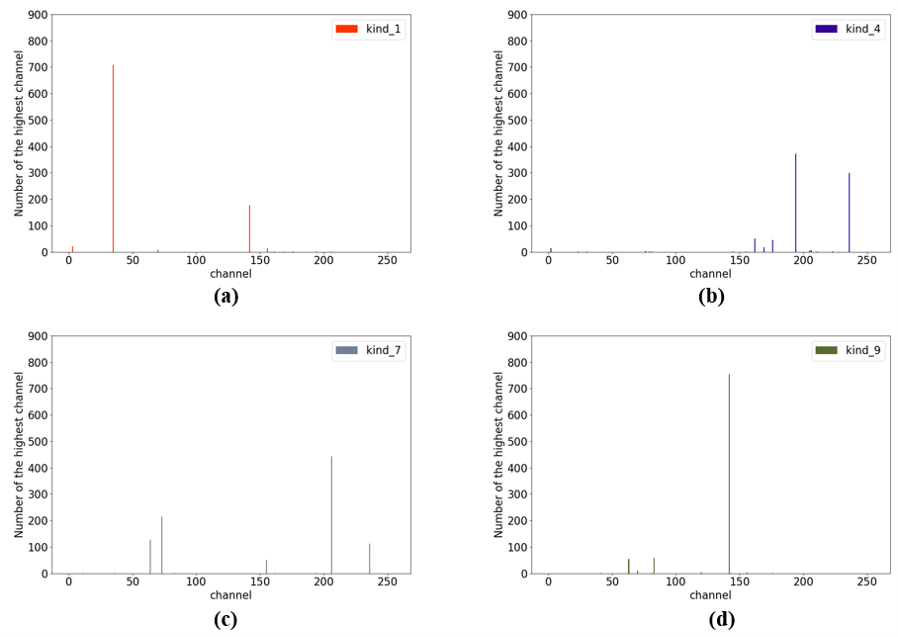}
\caption{The number of the channels with highest average activation value in trained ResNet-56 model's last layer for the images of four categories. We randomly choose four categories in CIFAR10. The channels with the highest activation value are different for different categories of images, and to successfully classified images in the same category, there are one or two fixed channels with the highest activation values. }
\label{fig2}
\end{center}
\end{figure*}

In this paper, we propose FocusedDropout, as a non-random dropout method to regularize CNN. Inspired by Network Dissection\cite{bau2017network}, a visualization method proposed by Hence, Bau et al., we first select the channel with the highest average activation value, and discard the units with an activation value below a threshold on this channel. Then, we can distinguish useful units from redundant units with the support of spatial invariance of CNN, by discarding all units having the same positions as previously discarded ones on the rest channels. Besides, increasing weight decay\cite{krogh1992simple} while using FocusedDropout can avoid overfitting and increase accuracy. As a result, the network focuses more on the units having the highest probability associated with the target, which is particularly effective for regularization. Extensive experimental results show that even a slight cost, 10\% of batches employ FocusedDropout, can produce a nice performance boost over the baselines on CIFAR10\cite{krizhevsky2009learning}, CIFAR100\cite{krizhevsky2009learning}, Tiny Imagenet\cite{TinyImagenet}, and has a good versatility for different CNN models including ResNet\cite{he2016deep}, DenseNet\cite{huang2017densely},VGGNet\cite{simonyan2014very} and Wide ResNet\cite{zagoruyko2016wide}.

\section{Motivation}
Existing non-random dropout methods judge units' influence by their ways and discard the units with greater influence, making network training more robust from the perspective of preventing overfitting. However, influential units may contain information that is conducive to classification, so strengthening the learning of this part of units can make networks focus more on classification target information, which inspires us to adopt a different dropout method: retaining influential units and discarding other units. There are two challenges: one is how to identify the target-related units; the other is how our method can prevent overfitting.

We select the target-related units based on Network Dissection\cite{bau2017network}, a visualization method for the interpretability of latent representations of CNN. In this study, after putting an image containing pixel-wise annotations for the semantic concepts into a trained network, the activation map of the target neuron is scaled up to the resolution of the ground truth segmentation mask. If the measurement result of alignment between the upsampled activation map and the ground truth segmentation mask is larger than a threshold, the neuron would be viewed as a visual detector for a specific semantic concept. From this work, we can conclude that the image's spatial information remains unchanged during training, and there are channels related to the classification target in the trained network. Then we need to find the target-related channels.

We conducted an exploratory experiment to find the target-related units. We first put the CIFAR10 dataset's validation set containing 10000 images into the vanilla ResNet-56 model trained with CIFAR10, and then recorded the channel in the last layer with highest activation value for every successfully classified image. Note that there are 9392 successfully classified images for 10 categories and the number of the channel in the last layer is 256. The result can be seen in Figure \ref{fig2}. The channels with the highest activation value are different for different categories of images, and for the same category of successfully classified images, there are only one or two fixed channels with the highest activation values. It showed that these channels are crucial to the success of the classification task. Thus we speculate these channels contain the features related to the image foreground. Besides, the channel with the highest average activation value has the highest probability of containing picture foreground information and the units with high activation value on this channel has the highest probability of representing foreground features. Therefore, we can take this channel as a reference, and combine spatial invariance to select other useful units on remaining channels.

Retaining only these units with larger weights may exacerbate overfitting intuitively. Thus we want to find a balance between strengthening the network's attention to the target and preventing the network from overfitting. To achieve this balance, we propose two countermeasures. First, we only randomly select 10\% of the batches to use FocusedDropout during every training epoch. Besides, we will magnify weight decay when using FocusedDropou. We found that these two measures can avoid overfitting effectively and improve the robustness of the network.

\section{Related work}

The regularization method has always been an effective way to improve the performance of neural networks, which involves two common regularization techniques: data augmentation and dropout\cite{srivastava2014dropout}. Unlike data augmentation, which expands the data space, dropout injects noise into feature space by randomly zeroing the activation function to avoid overfitting. Inspired by dropout, DropConnect\cite{wan2013regularization} drops the weights of the networks instead of activations to regularize large neural network models. Shakeout\cite{kang2016shakeout} randomly chooses to enhance or inverse the contributions of each unit to the next layer. Concrete Dropout\cite{gal2017concrete} utilizes concrete distributions to generate the dropout masks. Alpha-dropout\cite{ba2013adaptive} is designed for Scaled Exponential Linear Unit activation function. Variational dropout\cite{kingma2015variational} is proposed as a generalization of Gaussian dropout, but with a more flexibly parameterized posterior.

These strategies work well on a fully connected network, but their performance on CNN is unsatisfactory. The reason is that the correlation between the units on each channel in CNN is so strong that information can still flow through the networks despite dropout. To solve this problem, SpatialDropout\cite{tompson2015efficient} is proposed by randomly discarding the entire channels rather than individual activations. Channel-Drop\cite{huang2015channel} overcomes the drawbacks of local winner-take-all methods used in deep convolutional networks. Inspired by Cutout\cite{devries2017improved}, DropBlock\cite{ghiasi2018dropblock} randomly drops some contiguous regions of a feature map. With the continuous improvement of the network models, lots of regularization methods such as Stochastic Depth\cite{huang2016deep}, Shake-Shake\cite{gastaldi2017shake}, and ShakeDrop\cite{yamada2018shakedrop} have emerged for the specific CNN such as ResNet\cite{he2016deep}, ResNext\cite{xie2017aggregated}.

The above methods enhance the networks¡¯ ability by randomly discarding neurons or changing weight, but some approaches are focus on discarding specific units. Max-drop\cite{park2016analysis} drops the activations which have high values across the feature map or the channels. Targeted dropout\cite{gomez2018targeted} combines the concept of network cropping with dropout, which selects the least important weights in each round of training and discards the candidate weights to enhance network robustness. Guided Dropout\cite{keshari2019guided} defines the strength of each node and strengthens the training of weak nodes by discarding strong nodes during training. Weighted Channel Dropout\cite{hou2019weighted} aims to solve the overfitting in small data set training, which calculates the average activation value of each channel and retains the high-valued channels with higher possibility. AttentionDrop\cite{ouyang2019attentiondrop} drops features adaptively based on attention information. CamDrop\cite{wang2019camdrop} electively abandons some specific spatial regions in predominating visual patterns by considering the intensity of class activation mapping.

\section{Our Approach}

\begin{figure*}[ht]
\begin{center}
\includegraphics[width=0.8\linewidth]{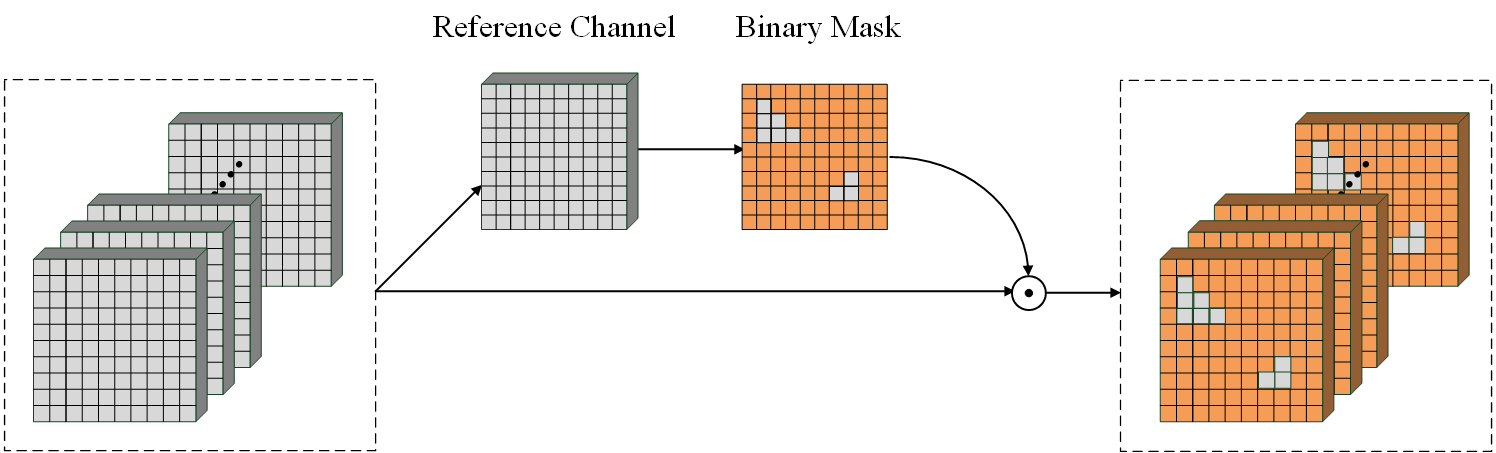}
\caption{Illustration of FocusedDropout. The channel with the highest activation value is selected as the reference channel to generate the binary mask. After covered by the mask, the units most likely to be related to the target will be retained, and the other will be discarded. The orange represents the neurons dropped by FocusedDropout. }
\label{fig3}
\end{center}
\end{figure*}

FocusedDropout is a highly targeted approach that makes the networks focus more on the foreground than background or noise. The main idea of FocusedDropout is keeping the units in the preferred locations related to the classification target and dropping the units in other locations. According to the spatial invariance of CNN, the locations of image's features are fixed during training so that different channels' units with the same spatial positions represent the same image features. Inspired by this phenomenon, FocusedDropout uses a binary mask to cover the target-independent units with the same positions on each channel, as demonstrated in Figure \ref{fig3}. The algorithm of FocusedDropout is illustrated in Algorithm \ref{al}. Next, we will present the details of FocusedDropout.

\begin{algorithm}[b]
\caption{FocusedDropout}
\label{al}
\LinesNumbered 
\KwIn{whole channels of the previous layer $C= [c_1,c_2,c_3,\dots, c_n]$), mode,}
\KwOut{$C^*$}
\If{mode = Inference, }{Return $C^*$ = $C$}
Calculate the average activation value for each $c_i$ as $w_i$, Get the $c_k$ with the max $w_i$ \;
Get the unit having the highest activation value as $c_k(\overline x,\overline y)$ \;  
Make the all-zero mask $m$ having the same size with $c_i$\;
\ForEach{$m(i,j)$ in $m$}{
\If{ $c_k(i,j)>random(0.6, 0.9) \cdot c_k(\overline x,\overline y)$ }{$m(i,j) =1$\;}
}
Return $C^* = C \odot m$
\end{algorithm}

\paragraph{Selecting the reference channel.}Each channel output by convolutional layers can be regarded as a set of features extracted from the image. Our goal is to find the channel with the highest possibility to obtain the features relevant to the target. We observed that the channel with the largest average activation value has the greatest effect on the result, and we consider it contains the most important features. Therefore, FocusedDropout uses Global Average Pooling to acquire the average activation value of each channel. To facilitate the presentation, we introduce the following concepts. $C=[c_1,c_2,..c_n]$ denotes the whole channels from the previous layer, $c_i$ denotes the single-channel i.e. the entire feature map, $c_i(x, y)$ denotes the activation value of an unit on $c_i$. So the average activation value of $c_i$ is computed as $$w_i = \frac{1}{w \times h}\sum_{x = 1} ^w \sum_{y = 1} ^h c_i(x,y) \eqno{(1)}$$

where $w$ and $h$ are the shared width and height of all channels. After getting all the weights, we choose the max one $c_k$ as the reference channel and $k$ is computed as $$k= \mathop{argmax}\limits_{i} w_i \eqno{(2)}$$

\begin{table*}[t]
\begin{center}
\begin{tabular}{ccccccc}
\toprule[1.2pt]
\midrule

Method   &\makecell[c]{ResNet20}   &\makecell[c]{ResNet56}   &\makecell[c]{ResNet110}   &\makecell[c]{VggNet19}   &\makecell[c]{DenseNet}  &\makecell[c]{WRN28}\\
\midrule

Baseline&91.48$\pm$0.21     &93.84$\pm$0.23     &94.32$\pm$0.16      &93.55$\pm$0.24       &95.30$\pm$0.12       &96.16$\pm$0.06\\

Dropout &91.51$\pm$0.31     &93.74$\pm$0.11      &94.52$\pm$0.23       &93.14$\pm$0.30      &95.32$\pm$0.08       &96.07$\pm$0.04\\

SpatialDropout &91.98$\pm$0.13       &94.28$\pm$0.20       &94.76$\pm$0.15      &93.19$\pm$0.12     &95.40$\pm$0.16      &96.13$\pm$0.02\\

DropBlock&92.15$\pm$0.14    &94.01$\pm$0.08     &94.92$\pm$0.07    &93.50$\pm$0.04     &95.40$\pm$0.31            &96.21$\pm$0.03\\
\midrule

FocusedDropout&{\bfseries 92.08$\pm$0.17}    &{\bfseries 94.67$\pm$0.07}     &{\bfseries 95.24$\pm$0.15}     &{\bfseries 93.74$\pm$0.19}       &{\bfseries 95.57$\pm$0.16}       &{\bfseries 96.48$\pm$0.13} \\
\midrule
\bottomrule[1.2pt]
\end{tabular}
\end{center}
\caption{Test accuracy (\%) on CIFAR10 dataset using CNN architectures of ResNet-20, ResNet-50, ResNet-110,VggNet-19, Densenet-100,Wide ResNet-28(Top accuracy are in bold). We report average over 3 runs.}
\label{tab1}
\end{table*}

\begin{table*}[htbp]
\begin{center}
\begin{tabular}{ccccccc}
\toprule[1.2pt]
\midrule

Method    &\makecell[c]{ResNet20}   &\makecell[c]{ResNet56}   &\makecell[c]{ResNet110}   &\makecell[c]{VggNet19}   &\makecell[c]{DenseNet}  &\makecell[c]{WRN28}\\
\midrule

Baseline&69.85$\pm$0.20     &73.71$\pm$0.28     &74.71$\pm$0.23      &73.14$\pm$0.16       &77.25$\pm$0.20       &81.27$\pm$0.20\\

Dropout &69.81$\pm$0.12     &73.81$\pm$0.27     &74.69$\pm$0.33     &73.01$\pm$0.21       &77.45$\pm$0.12       &81.21$\pm$0.11\\

SpatialDropout &69.71$\pm$0.05       &74.38$\pm$0.17       &74.76$\pm$0.12      &72.84$\pm$0.21     &77.97$\pm$0.16      &81.40$\pm$0.30\\

DropBlock&70.03$\pm$0.11    &73.92$\pm$0.10     &74.92$\pm$0.07    &73.01$\pm$0.04     &77.33$\pm$0.09            &81.15$\pm$0.09\\
\midrule

FocusedDropout&{\bfseries 70.27$\pm$0.06}    &{\bfseries 74.97$\pm$0.16}     &{\bfseries 76.43$\pm$0.17}     &{\bfseries 73.91$\pm$0.19}       &{\bfseries 78.35$\pm$0.16}       &{\bfseries 81.90$\pm$0.11} \\
\midrule
\bottomrule[1.2pt]
\end{tabular}
\end{center}
\caption{Test accuracy (\%) on CIFAR100 dataset. All settings are the same as on CIFAR10. We report average over 3 runs.}
\label{tab2}
\end{table*}

\paragraph{Finding the focused area.}After obtaining the reference channel, we need to get the location information of the target-related features from it. The units with high activation values have more influence to the task, so that we first find the highest activation value $c_k(\overline x,\overline y)$ on the reference channel as the criterion for setting the threshold, where $(\overline x,\overline y)$ is set as$$(\overline x,\overline y)=\mathop{argmax}\limits_{(x,y)} c_k(x,y) \eqno{(3)}$$
We take a random ratio $\gamma$ between 0.3 and 0.6 and set threshold as$$threshold=\gamma \cdot c^* (\overline x,\overline y) \eqno{(4)}$$
We set a random ratio rather than a certain one because different images have different numbers of features. Units with the activation value higher than the threshold are considered as target-related ones. The assembling of these units¡¯ positions is considered as the area related to the target, i.e., the focused area.

\paragraph{Generating the binary mask.} Although the focused area is obtained from the reference channel, units in the same area on the other channels are still related to the target due to the spatial invariance of CNN. In other words, different channels from the previous layer extract various features from the image, but the same position of these channels represents the same location of the image. Thus we need to retain each channel¡¯s units in the preferred area. To simplify the calculation, we use a binary mask multiplying all channels to achieve this step. The binary mask satisfys$$
m(i,j) =
\begin{cases}
0,    (i,j) \notin A \\
1,   (i,j) \in A
\end{cases}
\eqno{(5)}$$
where $m(i,j)$ represents the value of the binary mask, and $A$ represents the focused area. Only the units contained in the focused area can be reserved when the binary mask is multiplied by all channels.
\paragraph{Magnifying weight decay(MWD).}During training, we randomly choose 10\% of batches using FocusedDropout rather than every batch. Only retaining the high-weight units may aggravate the overfitting, so we increase weight decay when using FocusedDropout, which can limit the network's over attention to the classification target and inhibit the overfitting. It is a new training trick for dropout because dropout is often used in all batches of every epoch and weigh decay is fixed during training. We call the rate of batches applying FocusedDropout as $participation\_rate$. The computing resource and time consuming can be decreased obviously, and the model's performance can still be enhanced with this trick. In the testing phase, we retain all units as conventional dropout.

\section{Experiment}
In the following sections, we investigate the effectiveness of FocusedDropout mainly for image classification. We apply FocusedDropout to different networks for image classification on various kinds of datasets.

\paragraph{Implementation details} To evaluate our method's generalization, we apply FocusedDropout to three classification datasets: CIFAR10, CIFAR100, and Tiny Imagenet. Standard data augmentation scheme like flipping, random cropping are also incorporated. The hyper-parameters, including participation\_rate and $\gamma$, are set by cross-validation. For Wide ResNet-28, the learning rate is decayed by the factor of 0.1 at 60,120,160; for other networks, the learning rate is decayed by the factor of 0.1 at 150,225. Dropblock is applied to the output of first two groups. Dropout, SpatialDropout and FocusedDropout are applied to the output of penultimate group for fairness of comparison. The highest validation accuracy over the full training course is chosen as the result. All experiments are performed with Pytorch \cite{paszke2017automatic} on Tesla M40 GPUs.

\begin{figure*}[t]
\begin{center}
\includegraphics[width=0.9\linewidth]{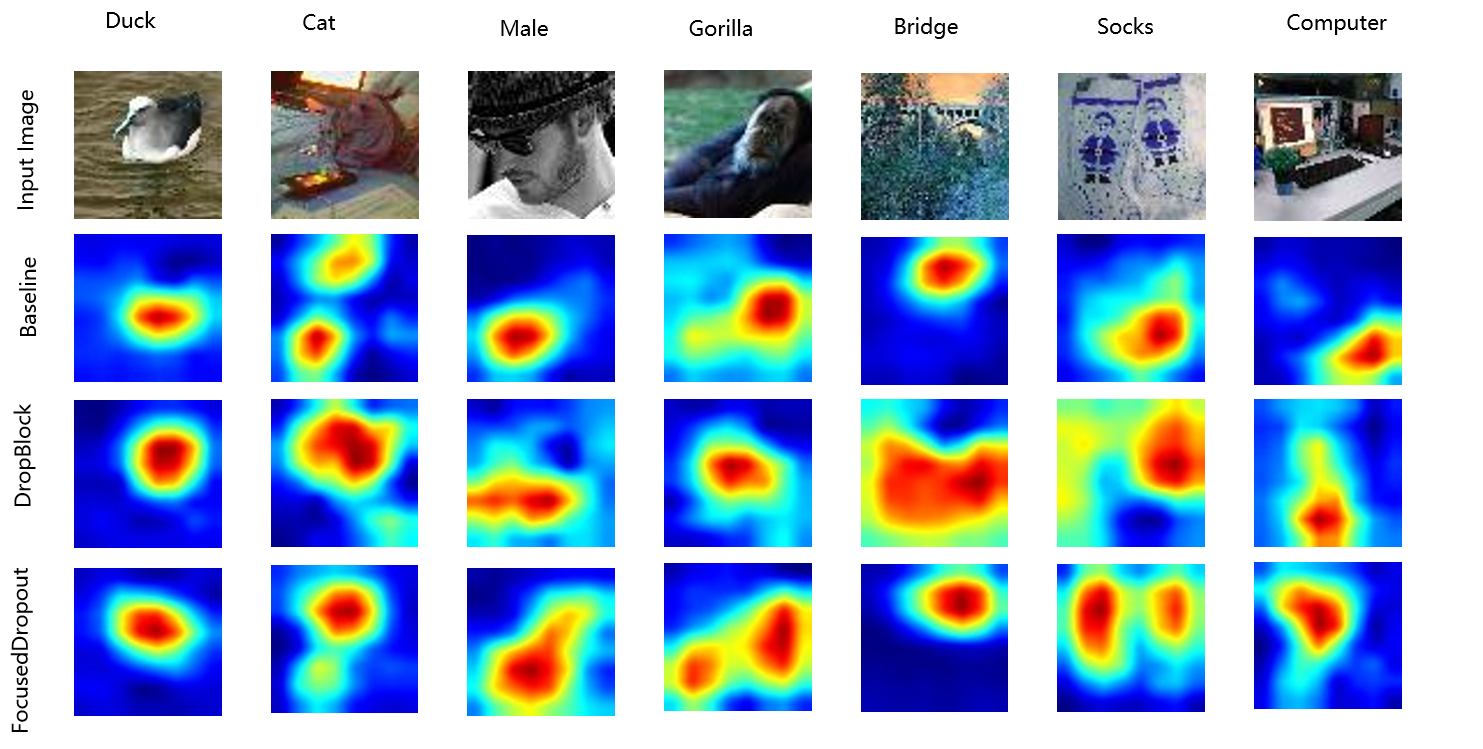}
\caption{Class activation mapping (CAM) \cite{zhou2016learning} for ResNet-110 model trained without any method, trained with DropBlock and trained with FocusedDropout. The model trained with our method can accurately identify the target's location and contour.}
\label{fig5}
\end{center}
\end{figure*}

\subsection{Evaluation of FocusedDropout}
\subsubsection{CIFAR10 and CIFAR100}
The CIFAR10 \cite{krizhevsky2009learning} dataset consists of 60,000 32$\times$32 color images of 10 classes, each with 6000 images including 50,00 training images and 10,00 test images. We adopt ResNet-20, ResNet-56, ResNet-110 \cite{he2016deep}, VGGNet-19 \cite{simonyan2014very}, DenseNet-100 \cite{huang2017densely} and Wide ResNet-28 \cite{zagoruyko2016wide} as the baselines to evaluate FocusedDropout's generalization for different structures of networks. We set $\gamma$ to a random value between 0.3 and 0.6, $participation\_rate$ = 0.1. As shown in Table \ref{tab1}, FocusedDropout achieves better improvement than other regularization methods in all networks. Generalization is an essential property of regularization methods. Experiments show that our method is suitable for networks with different layers, channels, and parameter spaces.

\begin{figure}[ht]
\begin{center}
\includegraphics[width=0.95\linewidth]{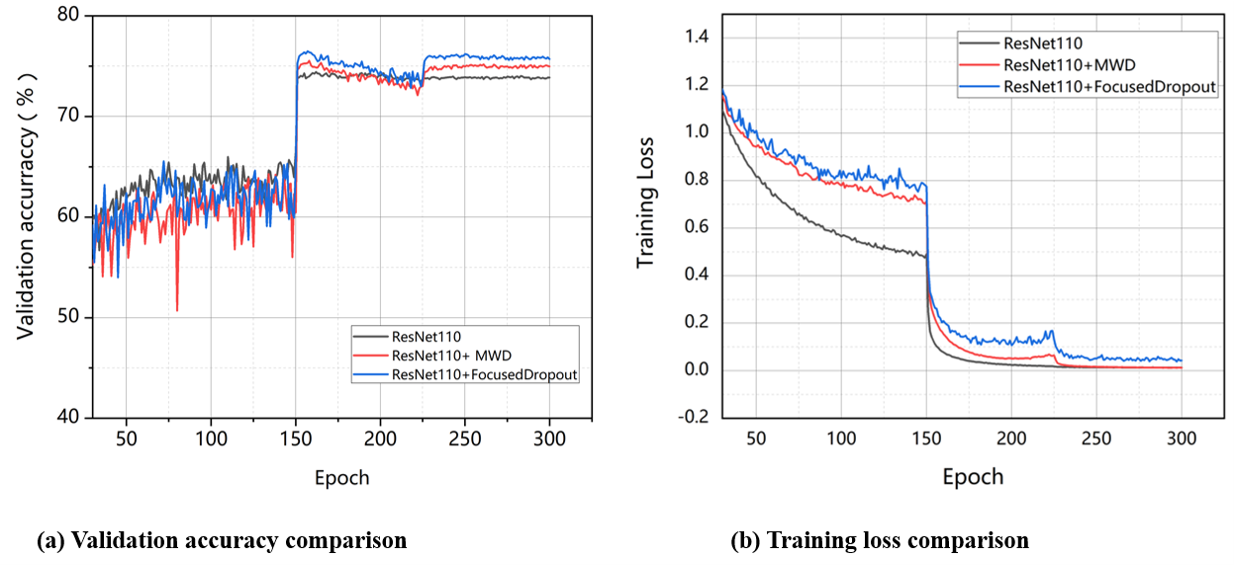}
\caption{Comparison of the validation accuracy curves and training loss curves of ResNet-110 on CIFAR100.}
\label{fig4}
\end{center}
\end{figure}

The CIFAR100 \cite{krizhevsky2009learning} dataset has the same number of images but 100 classes, which means the training of CIFAR100 is harder than CIFAR10 because of more image
types and less training data for each kinds. Results are summarized in Table \ref{tab2}. FocusedDropout still performs better than other dropout methods, and for ResNet-110, the promotion can even exceed 1.5\%. Validation accuracy and training loss comparison of ResNet-110 on CIFAR100 can be seen in Figure \ref{fig4}. The performance of the training networks with FocusedDropout and randomly Magnify Weight Decay(MWD) are worse than baseline before epoch 150, but when the learning rate decayed, the networks' performance would be much better than origin models. This shows that randomly MWD and FocusedDropout can effectively prevent overfitting.


\subsubsection{Tiny ImageNet}
Tiny ImageNet dataset \cite{TinyImagenet} is a subset of the ImageNet \cite{deng2009imagenet} dataset with 200 classes.  Each class has 500 training images, 50 validation images, and 50 test images. All images are with 64$\times$64 resolution. The test-set label is not publicly available, so we use the validation-set as a test-set for all the experiments on Tiny ImageNet following the common practice. We apply different dropout methods on ResNet-110 and compare the results with FocusedDropout. As shown in Table \ref{tab3}, FocusedDropout achieves a significant 2.16\% improvement over the base model and has a larger gain compared to other dropout methods.

To get an intuitive grasp of how FocusedDropout makes the network focus more on target, We use class activation mapping (CAM) \cite{zhou2016learning} to visualize the activation units of ResNet-110 trained by Tiny ImageNet as shown in Figure \ref{fig5}, from which we observe that the network trained with FocusedDropout can more accurately identify the target's location and contour, besides, the identification of the main target and the secondary target in the image is clearer. This proves that our approach enables the network to focus more on target. It also shows that making the network more goal-oriented helps improve the performance of the network.

\begin{table}[t]
\begin{center}
\begin{tabular}{lr}
\toprule[1.2pt]
\midrule
Metod    &Validation Accuracy(\%)  \\
\midrule
ResNet110 &\makecell[c]{62.42$\pm$0.25}\\

ResNet110+Dropout &\makecell[c]{62.32$\pm$0.05}\\

ResNet110+SpatialDropout &\makecell[c]{62.55$\pm$0.10}\\

ResNet110+DropBlock &\makecell[c]{63.13$\pm$0.29}\\

ResNet110+FocusedDropout &\makecell[c]{{\bfseries 64.58$\pm$0.16}}\\

\midrule
\bottomrule[1.2pt]
\end{tabular}
\end{center}
\caption{
The performance comparison on Tiny ImageNet dataset. The best accuracy is achieved by FocusedDropout. We report average over 3 runs.
}
\label{tab3}
\end{table}

\subsection{Ablation Study}

\begin{figure*}[t]
    \centering
    \subfigure[participation\_rate]{
    \begin{minipage}{4.5cm}
    \centering
        \includegraphics[width=5.3cm]{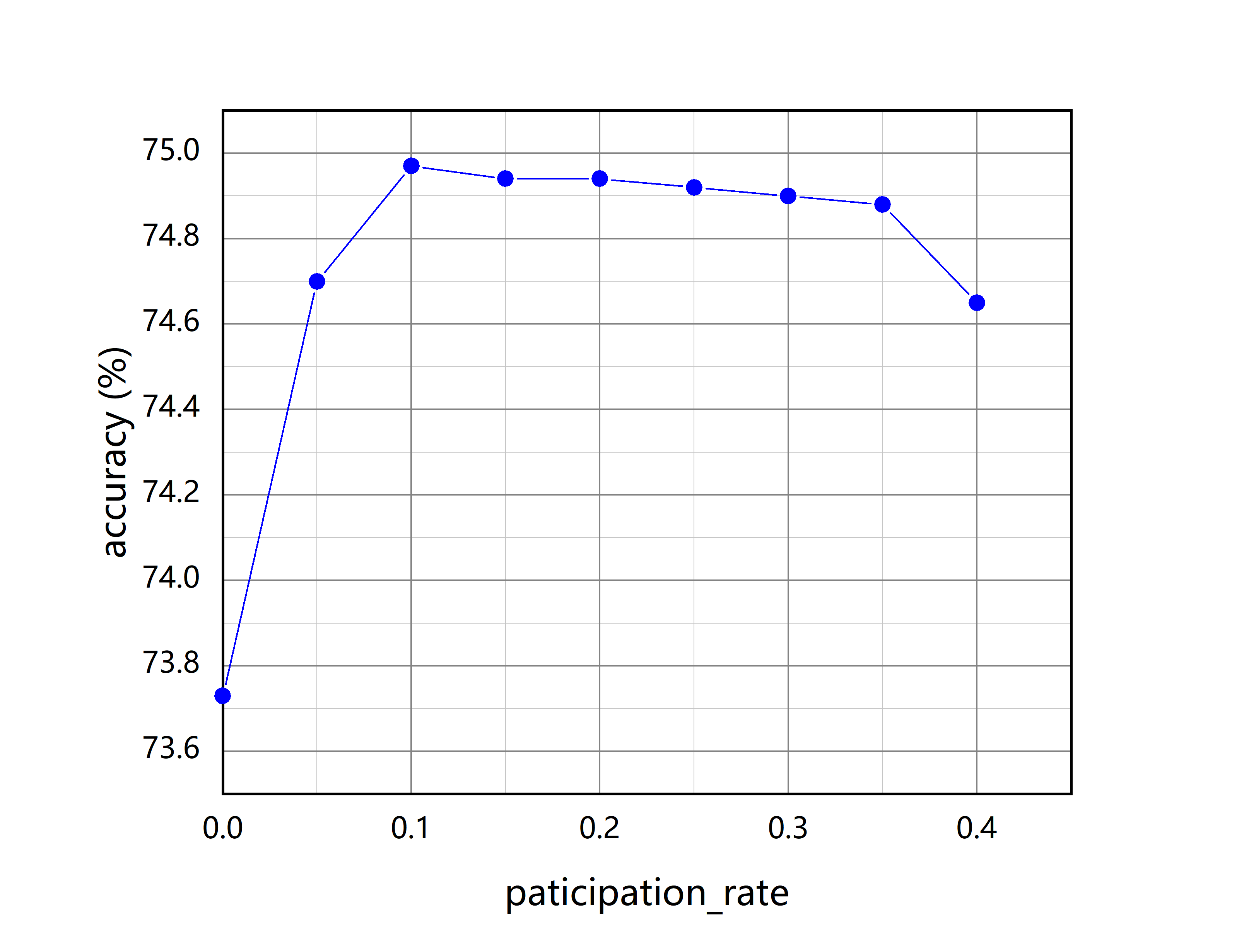}
    \end{minipage}%
    }
    \centering
    \subfigure[$\gamma$]{
    \begin{minipage}{4.5cm}
    \centering
        \includegraphics[width=5.3cm]{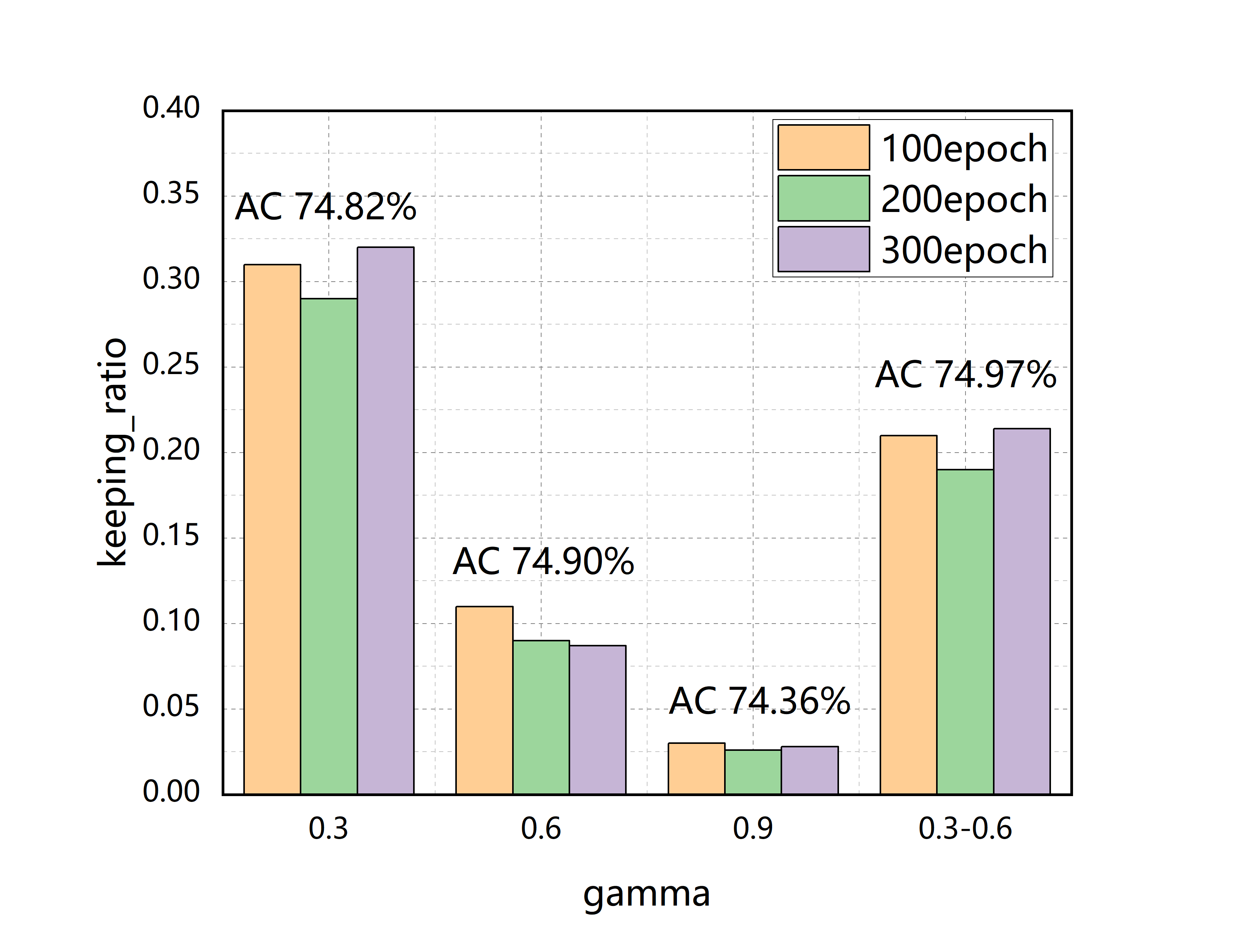}
    \end{minipage}%
    }
    \centering
    \subfigure[weight decay]{
    \begin{minipage}{4.5cm}
    \centering
        \includegraphics[width=5.3cm]{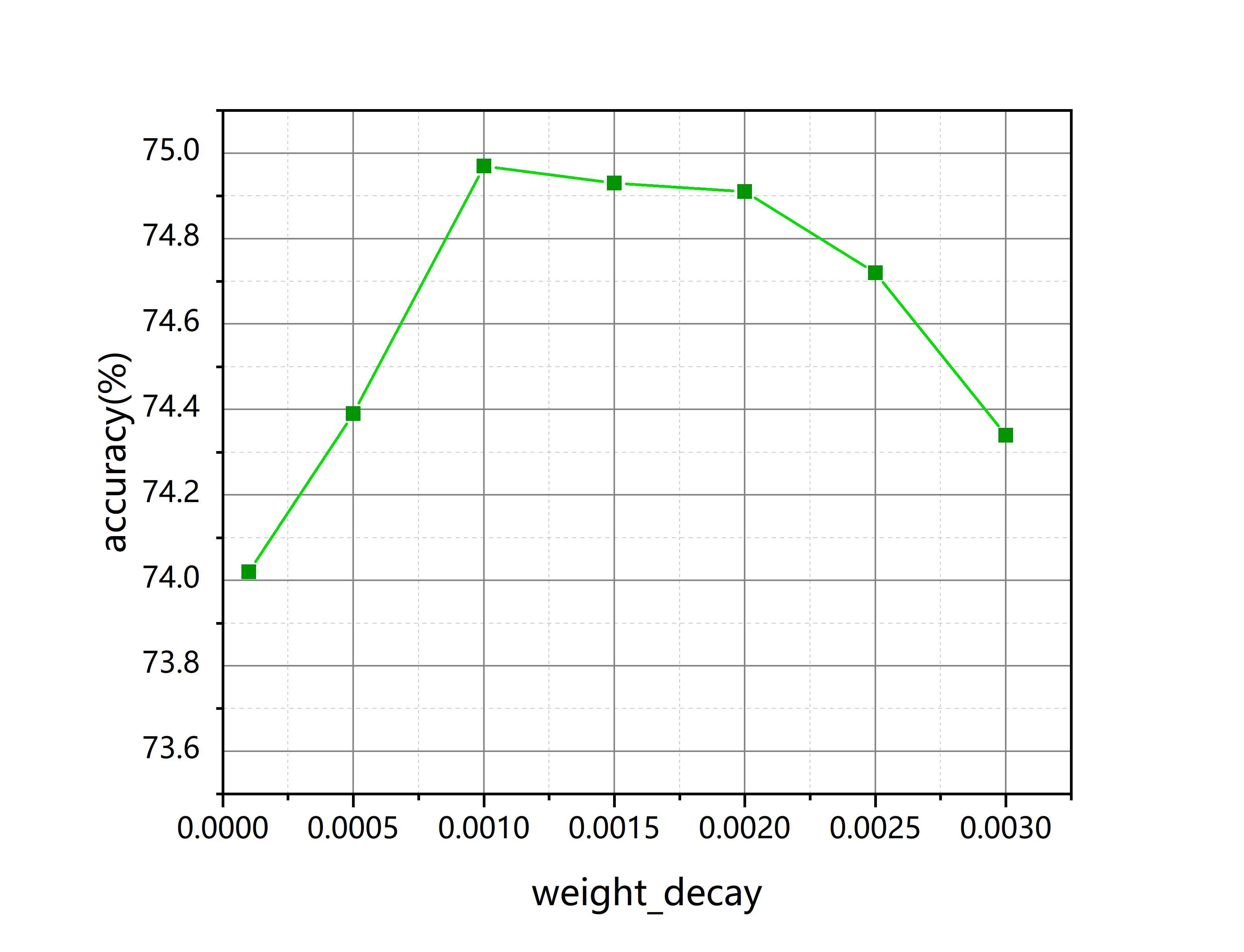}
    \end{minipage}%
    }
    \caption{CIFAR100 validation accuracy against particitpation\_rate, $\gamma$ and weight decay with ResNet56 model.}
    \label{fig6}
    \end{figure*}

In this subsection, we take extensive experiments to analyze the behaviors of FocusedDropout.\\

\subsubsection{Effect of the participation\_rate}
We explore the effect of the $participation\_rate$ on classification accuracy. Taking the performance of ResNet-56 on CIFAR100 as an example, we change the setting of $participation\_rate$ from 0 to 0.4.  As shown in Figure \ref{fig6} (a), with the increase of $participation\_rate$, the accuracy first rises and then decreases after reaching the highest when the $participation\_rate$ is 0.1. It indicates that increasing the $participation\_rate$ moderately can improve the network's performance, but excessive use of FousedDropout may lead to the deviation of the information learned by the network. Besides, the change is not apparent when $participation\_rate$ is below 0.3, so the parameter setting is not crucial for our method.

\subsubsection{Analysis on keeping\_ratio}
An obvious characteristic of FocusedDropout is that we do not directly set the $keeping\_ratio$ but choose a threshold to filter the units. We do some experiments to explore the $keeping\_ratio$ with a different threshold. Taking the performance of ResNet-56 on CIFAR100 as an example, we set $\gamma$ to 0.3, 0.6, 0.9, or a random number between 0.3 and 0.6. The ratio of the number of zeros on the binary mask to the total number of the mask's units is considered as the $keeping\_ratio$. The mask generated by each image is different, so we take the average of all $keeping\_ratio$ appeared in an epoch as the $keeping\_ratio$ of this epoch. As illustrated in Figure \ref{fig6} (b), the number of units discarded by FocusedDropout far exceeds the conventional dropout when $\gamma$ is greater than 0.3. Besides, Figure \ref{fig6} (b) shows how the threshold $\gamma$ affects accuracy with different values or ranges. When $\gamma$ is set to a random number, the performance is better than a fixed value. We consider different images have different features, so a random number is better than a fixed value.

\subsubsection{Exploration to weight decay}
We tested the effect of magnifying weight decay(MWD) when using FocuedDropout. As shown in Table 4, the performance of the network with higher weight decay is better. We consider that higher weight decay can limit the changing parameters, and the network will not focus too much on the target. Another notable phenomenon is with the same probability, only randomly expanding weight decay also improves the network's performance, but when combined with FocusedDropout, the performance improvement is even greater than the sum of the two. It demonstrates that FocusedDropout is a balance between making the network pay more attention to the target and preventing overfitting. Figure \ref{fig6} (c) shows with the increase of weight decay, the accuracy first rises and then decreases after reaching the highest when the weight decay is 0.001.

\subsubsection{Comparison with opposite approach}
We explored the performance of methods using the opposite approach to discard units. When training ResNet-56 on CIFAR100, we set the units included in the focused area on the binary mask to 1, and other positions to 0. Meanwhile, we keep the same parameter settings as FocusedDropout for training. The comparison can be seen in Table \ref{tab4}. Although the network's performance is still improved compared with baseline, it is not as good as FocusedDropout. We consider using the opposite way cannot make the network more focused on the target, and it is similar to SpatialDropout that achieves the effect of overcoming overfitting. Our method makes the network more focused on the target and also can prevent overfitting. This comparative test verifies that FocusedDropout improves network performance by enhancing network attention rather than the randomness.

\begin{table}[t]
\begin{center}
\begin{tabular}{lr}
\toprule[1.2pt]
\midrule
Model    & Accuracy(\%)  \\
\midrule

ResNet56&\makecell[c]{73.71$\pm$0.23}\\

+Randomly MWD&\makecell[c]{74.27$\pm$0.15}\\

+FocusedDropout without MWD&\makecell[c]{74.32$\pm$0.11}\\

+Opposite approach without MWD&\makecell[c]{73.96$\pm$0.08}\\

+Opposite approach&\makecell[c]{74.31$\pm$0.14}\\

+FocusedDropout&\makecell[c]{{\bfseries 74.97$\pm$0.16}}\\

\midrule
\bottomrule[1.2pt]
\end{tabular}
\end{center}
\caption{Comparisons to baseline, randomly magnifying weight decay, FocusedDropout without magnifying weight decay, opposite approach and FocusedDropout of test accuracy on CIFAR100 by using ResNet56.}
\label{tab4}
\end{table}

\subsection{Object Detection in PASCAL VOC}
In this subsection, we show FocusedDropout can also be applied for training object detector in Pascal VOC\cite{everingham2010pascal} dataset. RetinaNet\cite{lin2017focal} is used as the framework composed of a backbone network and two task-specific subnetworks for the experiments. The ResNet-50 backbone which is responsible for computing a convolutional feature map over an entire input image is initialized with ImageNet-pretrained model and then fine-tuned on Pascal VOC 2007 and 2012 trainval data. We apply FocusedDropout to ResNet-50 in ResNet-FPN.  The accuracy of ResNet-50 trained with FocusedDropout on ImageNet achieves 76.84\%, 0.52\% higher than the baseline(76.32\%). Models are evaluated on VOC 2007 test data using the mAP metric. We follow the fine-tuning strategy of the original method.

As shown in Table \ref{tab5}, the model pre-trained with FocusedDropout achieves 71.05\%, +0.91\% higher than the baseline performance. It shows that model trained with FocusedDropout can identify the target more easily and capture the position of the object more accurately.

\begin{table}[h]
\begin{center}
\begin{tabular}{lr}
\toprule[1.2pt]
\midrule
Method    &mAP(\%)  \\
\midrule

RetinaNet&\makecell[c]{70.14$\pm$0.17}\\

RetinaNet+FocusedDropout pretrained &\makecell[c]{{\bfseries 71.05$\pm$0.13}}\\

\midrule
\bottomrule[1.2pt]
\end{tabular}
\end{center}
\caption{Comparisons to baseline, randomly magnifying weight decay, FocusedDropout without magnifying weight decay, opposite approach and FocusedDropout of test accuracy on CIFAR100 by using ResNet56.}
\label{tab5}
\end{table}

\section{Conclusion}
In this work, we introduce a novel regularization method named FocusedDropout. Inspired by Network Dissection, we find that the high activation value units of CNN often corresponds to the classification target, so FocusedDropout first chooses the channel with the highest average activation value as the reference channel and find the preferred area from it, then only retain the units in this area for all channels due to the spatial invariance of CNN. Extensive experiments prove that FocusedDropout brings stable improvement to different datasets on various models. Besides, increasing weight decay when using FocusedDropout can prevent the network from overfitting.  The class activation mapping suggests the model can more accurately identify the target's location and contour regularized by FocusedDropout. We consider that FocusedDropout provides a new way to improve dropout: finding a balance between making the network focus on the target and preventing overfitting. For future work, we plan to apply FocusedDropout to other types of visual tasks, such as object detection.

{\small
\bibliographystyle{ieee_fullname}
\bibliography{re}

\begin{thebibliography}{10}\itemsep=-1pt

\bibitem{TinyImagenet}
\url{https://tiny-imagenet.herokuapp.com}.

\bibitem{ba2013adaptive}
Jimmy Ba and Brendan Frey.
\newblock Adaptive dropout for training deep neural networks.
\newblock In {\em Advances in Neural Information Processing Systems}, pages
  3084--3092, 2013.

\bibitem{bau2017network}
David Bau, Bolei Zhou, Aditya Khosla, Aude Oliva, and Antonio Torralba.
\newblock Network dissection: Quantifying interpretability of deep visual
  representations.
\newblock In {\em Proceedings of the IEEE Conference on Computer Vision and
  Pattern Recognition}, pages 6541--6549, 2017.

\bibitem{deng2009imagenet}
Jia Deng, Wei Dong, Richard Socher, Li-Jia Li, Kai Li, and Li Fei-Fei.
\newblock Imagenet: A large-scale hierarchical image database.
\newblock In {\em 2009 IEEE conference on computer vision and pattern
  recognition}, pages 248--255. Ieee, 2009.

\bibitem{devries2017improved}
Terrance DeVries and Graham~W Taylor.
\newblock Improved regularization of convolutional neural networks with cutout.
\newblock {\em arXiv preprint arXiv:1708.04552}, 2017.

\bibitem{everingham2010pascal}
Mark Everingham, Luc Van~Gool, Christopher~KI Williams, John Winn, and Andrew
  Zisserman.
\newblock The pascal visual object classes (voc) challenge.
\newblock {\em International journal of computer vision}, 88(2):303--338, 2010.

\bibitem{gal2017concrete}
Yarin Gal, Jiri Hron, and Alex Kendall.
\newblock Concrete dropout.
\newblock In {\em Advances in Neural Information Processing Systems}, pages
  3581--3590, 2017.

\bibitem{gastaldi2017shake}
Xavier Gastaldi.
\newblock Shake-shake regularization.
\newblock {\em arXiv preprint arXiv:1705.07485}, 2017.

\bibitem{ghiasi2018dropblock}
Golnaz Ghiasi, Tsung-Yi Lin, and Quoc~V Le.
\newblock Dropblock: A regularization method for convolutional networks.
\newblock In {\em Advances in Neural Information Processing Systems}, pages
  10727--10737, 2018.

\bibitem{gomez2018targeted}
Aidan~N Gomez, Ivan Zhang, Kevin Swersky, Yarin Gal, and Geoffrey~E Hinton.
\newblock Targeted dropout.
\newblock 2018.

\bibitem{he2016deep}
Kaiming He, Xiangyu Zhang, Shaoqing Ren, and Jian Sun.
\newblock Deep residual learning for image recognition.
\newblock In {\em Proceedings of the IEEE conference on computer vision and
  pattern recognition}, pages 770--778, 2016.

\bibitem{hou2019weighted}
Saihui Hou and Zilei Wang.
\newblock Weighted channel dropout for regularization of deep convolutional
  neural network.
\newblock AAAI, 2019.

\bibitem{huang2017densely}
Gao Huang, Zhuang Liu, Laurens Van Der~Maaten, and Kilian~Q Weinberger.
\newblock Densely connected convolutional networks.
\newblock In {\em Proceedings of the IEEE conference on computer vision and
  pattern recognition}, pages 4700--4708, 2017.

\bibitem{huang2016deep}
Gao Huang, Yu Sun, Zhuang Liu, Daniel Sedra, and Kilian~Q Weinberger.
\newblock Deep networks with stochastic depth.
\newblock In {\em European conference on computer vision}, pages 646--661.
  Springer, 2016.

\bibitem{huang2015channel}
Yuchi Huang, Xiuyu Sun, Ming Lu, and Ming Xu.
\newblock Channel-max, channel-drop and stochastic max-pooling.
\newblock In {\em Proceedings of the IEEE Conference on Computer Vision and
  Pattern Recognition Workshops}, pages 9--17, 2015.

\bibitem{kang2016shakeout}
Guoliang Kang, Jun Li, and Dacheng Tao.
\newblock Shakeout: A new regularized deep neural network training scheme.
\newblock In {\em Thirtieth AAAI Conference on Artificial Intelligence}, 2016.

\bibitem{keshari2019guided}
Rohit Keshari, Richa Singh, and Mayank Vatsa.
\newblock Guided dropout.
\newblock In {\em Proceedings of the AAAI Conference on Artificial
  Intelligence}, volume~33, pages 4065--4072, 2019.

\bibitem{kingma2015variational}
Durk~P Kingma, Tim Salimans, and Max Welling.
\newblock Variational dropout and the local reparameterization trick.
\newblock In {\em Advances in Neural Information Processing Systems}, pages
  2575--2583, 2015.

\bibitem{krizhevsky2009learning}
Alex Krizhevsky, Geoffrey Hinton, et~al.
\newblock Learning multiple layers of features from tiny images.
\newblock Technical report, Citeseer, 2009.

\bibitem{krogh1992simple}
Anders Krogh and John~A Hertz.
\newblock A simple weight decay can improve generalization.
\newblock In {\em Advances in neural information processing systems}, pages
  950--957, 1992.

\bibitem{lin2017focal}
Tsung-Yi Lin, Priya Goyal, Ross Girshick, Kaiming He, and Piotr Doll{\'a}r.
\newblock Focal loss for dense object detection.
\newblock In {\em Proceedings of the IEEE international conference on computer
  vision}, pages 2980--2988, 2017.

\bibitem{ouyang2019attentiondrop}
Zhihao Ouyang, Yan Feng, Zihao He, Tianbo Hao, Tao Dai, and Shu-Tao Xia.
\newblock Attentiondrop for convolutional neural networks.
\newblock In {\em 2019 IEEE International Conference on Multimedia and Expo
  (ICME)}, pages 1342--1347. IEEE, 2019.

\bibitem{park2016analysis}
Sungheon Park and Nojun Kwak.
\newblock Analysis on the dropout effect in convolutional neural networks.
\newblock In {\em Asian Conference on Computer Vision}, pages 189--204.
  Springer, 2016.

\bibitem{paszke2017automatic}
Adam Paszke, Sam Gross, Soumith Chintala, Gregory Chanan, Edward Yang, Zachary
  DeVito, Zeming Lin, Alban Desmaison, Luca Antiga, and Adam Lerer.
\newblock Automatic differentiation in pytorch.
\newblock 2017.

\bibitem{simonyan2014very}
Karen Simonyan and Andrew Zisserman.
\newblock Very deep convolutional networks for large-scale image recognition.
\newblock {\em arXiv preprint arXiv:1409.1556}, 2014.

\bibitem{srivastava2014dropout}
Nitish Srivastava, Geoffrey Hinton, Alex Krizhevsky, Ilya Sutskever, and Ruslan
  Salakhutdinov.
\newblock Dropout: a simple way to prevent neural networks from overfitting.
\newblock {\em The journal of machine learning research}, 15(1):1929--1958,
  2014.

\bibitem{tompson2015efficient}
Jonathan Tompson, Ross Goroshin, Arjun Jain, Yann LeCun, and Christoph Bregler.
\newblock Efficient object localization using convolutional networks.
\newblock In {\em Proceedings of the IEEE Conference on Computer Vision and
  Pattern Recognition}, pages 648--656, 2015.

\bibitem{wan2013regularization}
Li Wan, Matthew Zeiler, Sixin Zhang, Yann Le~Cun, and Rob Fergus.
\newblock Regularization of neural networks using dropconnect.
\newblock In {\em International conference on machine learning}, pages
  1058--1066, 2013.

\bibitem{wang2019camdrop}
Hongjun Wang, Guangrun Wang, Guanbin Li, and Liang Lin.
\newblock Camdrop: A new explanation of dropout and a guided regularization
  method for deep neural networks.
\newblock In {\em Proceedings of the 28th ACM International Conference on
  Information and Knowledge Management}, pages 1141--1149, 2019.

\bibitem{xie2017aggregated}
Saining Xie, Ross Girshick, Piotr Doll{\'a}r, Zhuowen Tu, and Kaiming He.
\newblock Aggregated residual transformations for deep neural networks.
\newblock In {\em Proceedings of the IEEE conference on computer vision and
  pattern recognition}, pages 1492--1500, 2017.

\bibitem{yamada2018shakedrop}
Yoshihiro Yamada, Masakazu Iwamura, Takuya Akiba, and Koichi Kise.
\newblock Shakedrop regularization for deep residual learning.
\newblock {\em arXiv preprint arXiv:1802.02375}, 2018.

\bibitem{zagoruyko2016wide}
Sergey Zagoruyko and Nikos Komodakis.
\newblock Wide residual networks.
\newblock {\em arXiv preprint arXiv:1605.07146}, 2016.

\bibitem{zhou2016learning}
Bolei Zhou, Aditya Khosla, Agata Lapedriza, Aude Oliva, and Antonio Torralba.
\newblock Learning deep features for discriminative localization.
\newblock In {\em Proceedings of the IEEE conference on computer vision and
  pattern recognition}, pages 2921--2929, 2016.

\end{thebibliography}
}

\end{document}